\documentclass[letterpaper, 10 pt, conference]{ieeeconf}  

\IEEEoverridecommandlockouts                              

\usepackage{graphics} 
\usepackage{subfigure}
\usepackage{epsfig} 
\usepackage{multicol}
\usepackage{amssymb}  
\usepackage{algorithm}
\usepackage{algorithmicx}
\usepackage{algpseudocode}
\usepackage{epstopdf}
\usepackage{multirow}
\usepackage{cite}
\usepackage{amsmath}
\usepackage{hyperref}
\usepackage{array}
\usepackage{balance}
\usepackage[flushleft]{threeparttable}
\newtheorem{theorem}{Theorem}

\title{\LARGE \bf
	Algorithm Design and Integration for a Robotic Apple Harvesting System
}

\author{Kaixiang Zhang$^{*}$,
	Kyle Lammers$^{*}$,
	Pengyu Chu,
	Nathan Dickinson,
	Zhaojian Li$^{**}$,
	and Renfu Lu 
	\thanks{This work is supported by the USDA-ARS inhouse project 5050-43640-003-00D.}
	\thanks{Kaixiang Zhang, Kyle Lammers, Pengyu Chu, and Zhaojian Li are with the Department of Mechanical Engineering, Michigan State University, East Lansing, MI 48824, USA (e-mail: zhangk64@msu.edu; lammer18@msu.edu; chupengy@msu.edu; lizhaoj1@egr.msu.edu).}
	\thanks{Nathan Dickinson is with the Department of Biosystems and Agricultural Engineering, Michigan State University, East Lansing, MI 48824, USA (e-mail: dicki112@msu.edu).}
	\thanks{Renfu Lu is with the United States Department of Agriculture Agricultural Research Service, East Lansing, MI 48824, USA (e-mail: renfu.lu@usda.gov).}
	\thanks{* The authors contribute equally to this paper.}
	\thanks{** Zhaojian Li is the corresponding author.}
}

\begin{document}
	
	\maketitle
	\thispagestyle{empty}
	\pagestyle{empty}
	
	\begin{abstract}
		Due to labor shortage and rising labor cost for the apple industry, there is an urgent need for the development of robotic systems to efficiently and autonomously harvest apples. 
		In this paper, we present a system overview and algorithm design of our recently developed robotic apple harvester prototype. Our robotic system is enabled by the close integration of several core modules, including visual perception, planning, and control. This paper covers the main methods and advancements in deep learning-based multi-view fruit detection and localization, unified picking and dropping planning, and dexterous manipulation control. Indoor and field experiments were conducted to evaluate the performance of the developed system, which achieved an average picking rate of 3.6 seconds per apple. This is a significant improvement over other reported apple harvesting robots with a picking rate in the range of 7-10 seconds per apple. The current prototype shows promising performance towards further development of efficient and automated apple harvesting technology. Finally, limitations of the current system and future work are discussed.
		
	\end{abstract}

	\section{Introduction}
	
	The apple industry relies heavily on manual labor. For instance, in the United States alone, 
	it is estimated that the seasonal labor force needed for apple harvesting is more than 10 million worker hours each year, attributing to about 15\% of the total production costs \cite{gallardo2012}. The growing labor shortage and increased labor cost have thus become major concerns for the long-term sustainability and profitability of the apple industry. In the meantime, the past decade has seen great transitions in apple production systems; traditional unstructured orchards have been replaced with  high-density orchard systems where trees are smaller and more uniformly structured (i.e., v-trellis, vertical fruiting wall, etc.). These modern tree structures can greatly facilitate orchard automation, and thus there has been a renewed interest in pursuing  robotic harvesting  as a promising solution to reduce the harvesting cost and dependence on manual labor. 
	
	Over the past few years, several robotic systems have been designed to autonomously harvest different horticultural crops, including sweet pepper\cite{LehnertRAL2017}, strawberry\cite{Xiong2020}, apple \cite{silwal2017}, and kiwifruit\cite{Williams2020}. For apple harvesting, the automation system designs can be mainly grouped into two categories. The first category is the \textit{shake-and-catch} harvesting \cite{ZhangASABE2020}, where vibrations are applied to the tree trunk and/or branches to detach the fruits. Although the shake-and-catch harvesting systems are efficient in 
	detaching fruits from trees, they often result in a high rate of apple bruising that is not acceptable for fresh market. The other category is the \textit{fruit-by-fruit harvesting} where  manipulators are used to pick fruits in a controlled manner, and thus can substantially reduce fruit damage. However, designing such systems with high picking efficiency and practical viability presents a great challenge. 
	
	So far, several fruit-by-fruit robotic apple harvesting systems have been developed \cite{baeten2008,silwal2017,Hohimer2019,Zhang2021MECH,Bulanon2021}. For instance, Baeton et al.   combines a 7 degree-of-freedom (DOF) industrial manipulator with a vacuum activated, funnel shaped gripper for apple detachment, and the harvesting cycle time is 8-10 s/fruit \cite{baeten2008}. In \cite{silwal2017}, both hardware and software designs of an apple harvester are presented. Field tests conducted on a v-trellis orchard show that this system is able to pick 84\% of 150 apples attempted with the overall harvesting time being 7.6 s/fruit. In \cite{Hohimer2019}, Hohimer et al. developed a harvesting robot based on a pneumatic soft-robotic end-effector, and the average time that the system takes from apple detachment to transported to storage bin is 7.3 s/fruit.
	Despite the aforementioned progresses, the low picking efficiencies of existing systems are still unsatisfactory for  their practical use  in the real orchard environment \cite{lu2017innovative}.
	
	Towards the goal of developing a practically and economically viable robotic harvesting system, we have been developing an efficient automated apple harvesting system over the past three years. Tests in orchard field and indoor simulated orchard environment demonstrated a promising picking rate of $\sim$3.6 s/fruit, a significant improvement over the existing systems reviewed above. While the mechanical and preliminary control designs have been reported in \cite{Zhang2021MECH}, this paper presents the algorithm design and integration of the developed system, focusing on three major modules -- perception, planning, and control -- where several advancements have been made. 
	First, we develop a deep learning-based multi-view fruit detection and localization framework by fusing two RGB-D sensors facing different angles. Compared to the single camera-based algorithm we developed earlier \cite{Chu2020PRL}, this multi-view fusion offers enhanced performance in both detection and localization. Second, a unified planning algorithm that simultaneously optimizes picking sequence and dropping spots is developed, which significantly improves harvesting efficiency. Lastly, a computationally-efficient nonlinear controller is synthesized to enable accurate and smooth manipulator movement. Experiments in both an indoor simulated orchard environment and a real orchard field were conducted to illustrate the performance of the integrated system.
	
	
	The remainder of this paper is organized as follows. 
	Section \ref{sec_systemOverview} provides an overview of  the developed robotic apple harvesting system. The algorithm designs of perception, planning and control modules are detailed in Section \ref{sec_algorithm}, and  experiment results are discussed in Section \ref{sec_perfEva}. Finally, conclusions are drawn in Section \ref{sec_conclusion}.

	\section{System Overview} \label{sec_systemOverview}
\begin{figure}[!h]
		\centering
		\includegraphics[width=0.9\linewidth]{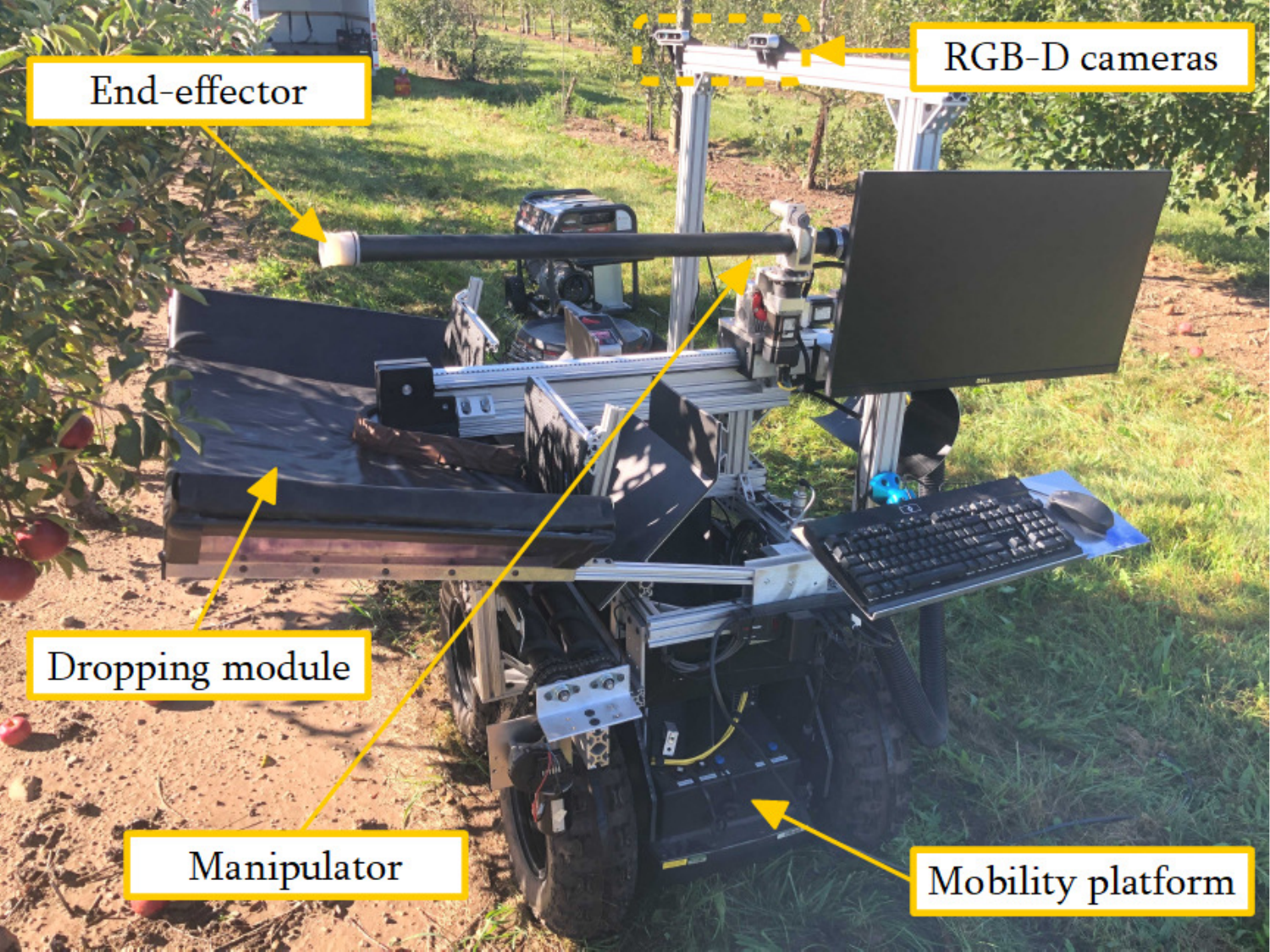}
		\caption{The developed robotic apple harvesting prototype.}\label{fig_appleRob}
	\end{figure}

	The developed robotic apple harvesting system is shown in Fig. \ref{fig_appleRob}, which is comprised of four primary hardware components: a perception module consisting of two Intel RealSense D435i RGB-D cameras, a 3-DOF manipulator, a vacuum-based end-effector, and a dropping module. All components are affixed to a Segway mobility platform for ease of movement in orchard. 
	The RGB-D cameras, the manipulator, and all communication devices are connected to an industrial computer (Xeon E2176G CPU and 64 GB RAM) resided in the mobility platform. The robot operating system (ROS) is used to fully integrate the entire software  and facilitate the communication and control of different components. 
	
	\subsection{Hardware Design} \label{sub_hardware}
	For automated apple harvesting, the first and foremost task is orchard perception, which detects and localizes the fruits to guide robotic manipulations. 
	Different from  existing works (e.g., \cite{baeten2008,Bulanon2021}) that attach the camera to the manipulator or the end-effector, the RGB-D cameras are installed on the Segway mobility platform to provide a global view of the scene, facilitating the use of multiple manipulator arms planned in our future versions. Moreover, the multi-camera setup is introduced to provide multi-view sensing from different perspectives, which is intended to achieve enhanced perception accuracy and robustness through sensor fusion to alleviate the impact of occlusions and challenging lighting conditions. 
	
	To efficiently approach the target fruits, a 3-DOF manipulator with simple and compact mechanical structure is designed and assembled. Specifically, the manipulator is comprised of one prismatic joint and two revolute joints. The two revolute joints are linked using an $L$-shaped aluminum plate, which creates a pan-and-tilt module. The prismatic joint is assembled as the base of the pan-and-tilt module to extend the depth of the manipulator's workspace. A hollow aluminum link is installed on the pan-and-tilt module to ensure that the end-effector can reach the apple locations, and it also acts as a vacuum tube for grasping fruits in the harvesting process. Instead of relying on a hybrid pneumatic/motor actuation mechanism in our previous design \cite{Zhang2021MECH}, all the joints of the current manipulator are driven by servo motors, which not only reduces actuation complexity but also facilitates integrated control scheme design.
	
	In our system, a vacuum-based end-effector is designed to grasp and detach fruits. A soft silicone vacuum cup is attached to the front end of the aluminum tube. The vacuum cup with a special geometric configuration has shown satisfactory performance in conforming to various apple contours \cite{Dickinson2022}. Meanwhile, the rear end of the aluminum tube is connected to an electric powered wet/dry vacuum via a flexible and expandable tube. The vacuum-based end-effector can reduce potential damage to fruits. Moreover, if the manipulator does not reach the apple accurately, the vacuum-based end-effector can tolerate some approaching inaccuracies since it can attract the fruit within a certain distance (about 1.5 cm in our current prototype) when adequate vacuum flow is provided. 
	
	For ease of collecting and transporting picked apples, a dropping module is assembled and affixed to the mobility platform. The base of the dropping module is a rectangular aluminum plate with a foam cushion covering.
	The manipulator can stop at any spots above the dropping module and then release the harvested fruit, thus reducing the harvesting cycle time. After the apple have fallen on the sloped surface of the dropping module, it rolls down to the rear end of the dropping module where a screw-driven conveyer is installed to transport the apple to a bin \cite{zhang2021PBT}.
	
	\subsection{Software Design}
\begin{figure}[!h]
		\centering
		\includegraphics[width=0.85\linewidth]{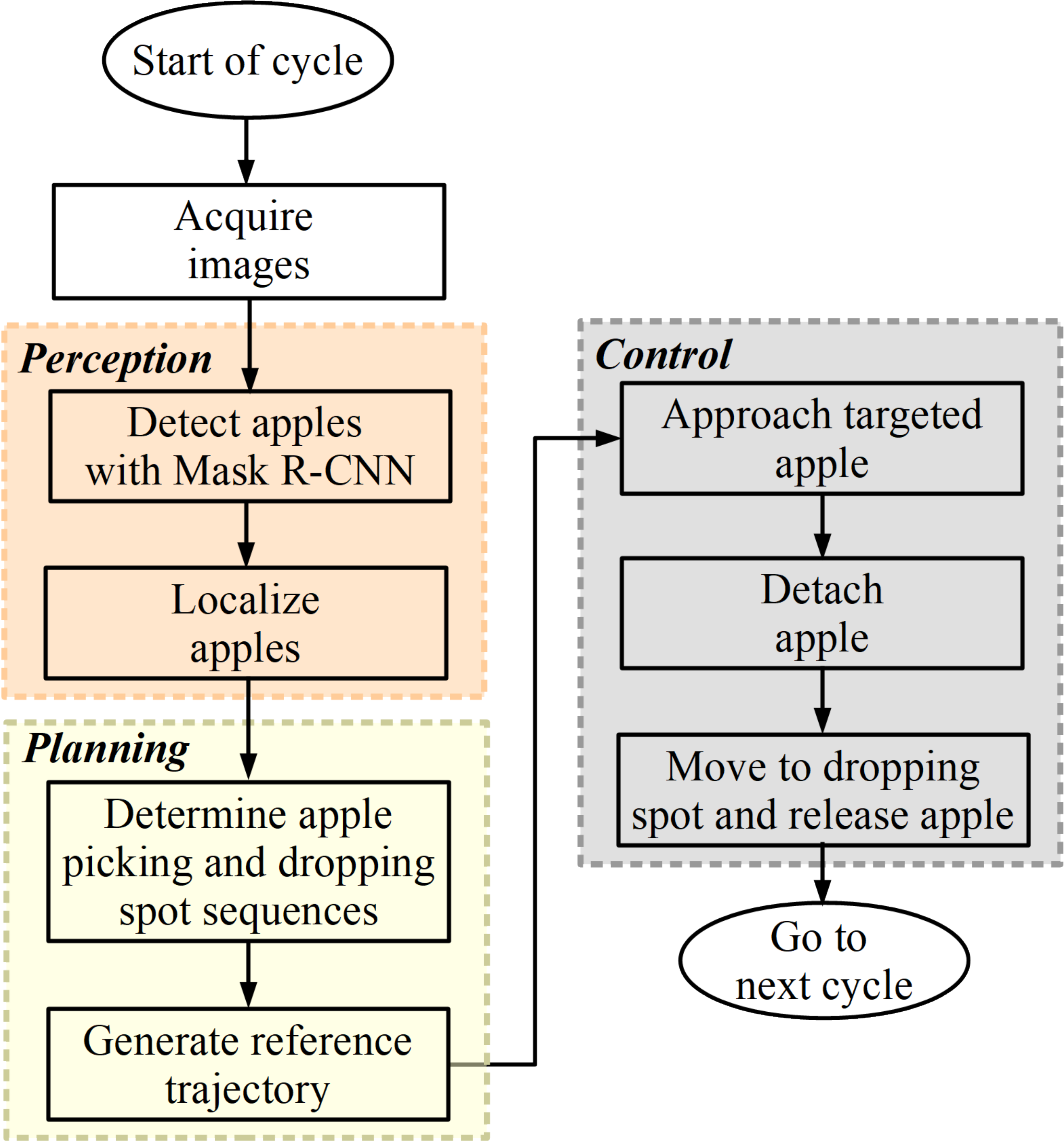}
		\caption{Algorithm flowchart in an apple harvesting cycle.}\label{fig_flowDiagram}
	\end{figure}

	The software suite is designed and integrated in the ROS framework. Different software components are primarily communicated via custom messages sent through ROS actions and services. Fig. \ref{fig_flowDiagram} shows the main algorithm flow of the software system during an apple harvesting cycle. The algorithm structure mainly consists of three modules: perception, planning, and control. The logic flow of the apple harvesting cycle is detailed in the following.
	
	At the beginning of each harvesting cycle, the RGB-D cameras are triggered to acquire images. With the obtained image information, the perception algorithm (Section~\ref{subsec_visualSensing}) is used to detect and localize the fruits within the system's workspace. A list of 3D apple locations are then generated and subsequently transformed into the 3D positions expressed in the coordinate frame of the manipulator. 
	Based on the apple location list, the planning algorithm is utilized to optimize the apple picking sequence and its corresponding dropping spots (Section~\ref{subsec_planning}). The detected apples will be chosen as the targeted fruits by following the planned picking sequence, and a reference trajectory will be generated to guide the motion of the manipulator. The target apple location and its corresponding reference trajectory are passed onto the control module, which then actuates the manipulator to follow the reference trajectory to reach the fruit. Once the fruit is successfully attached to the end-effector (detected by a pressure sensor mounted inside the tube), the rotation mechanism is triggered to rotate the whole aluminum tube by a certain angle to detach the apple (Section~\ref{subsec_control}). Finally, the manipulator returns to a dropping spot and releases the fruit. It is apparent that the software design of our robotic system requires multi-disciplinary advances to enable various synergistic functionalities and coordination for achieving reliable automated apple harvesting. The next section describes each of the software components in more details. 
	
	\section{Algorithm Design and Integration} \label{sec_algorithm}
	In this section, we describe our software components on perception, planning, and control in details.

	\subsection{Multi-view fusion for robust detection and localization} \label{subsec_visualSensing}
	
	One of the key tasks in robotic apple harvesting is fruit detection and localization, where the former is to segment apples from the background areas whereas the latter subsequently calculates the 3D positions of the detected apples. In our preliminary work, a network with Mask R-CNN backbone and  a suppression end was developed in \cite{Chu2020PRL} using a single RGB-D camera. In this new version, we extend the perception system to systematically fuse two RGB-D cameras to enhance the detection performance. This is motivated to address the two major challenges in orchard perception as identified in our previous filed tests: leaf/branch occlusion and varying lighting conditions. Exploiting multiple cameras from different views can alleviate the impact of occlusion and challenging lighting conditions as the two cameras can provide complementary views for enhanced performance.
	\begin{figure}[!t]
		\centering
		\includegraphics[width=0.98\linewidth]{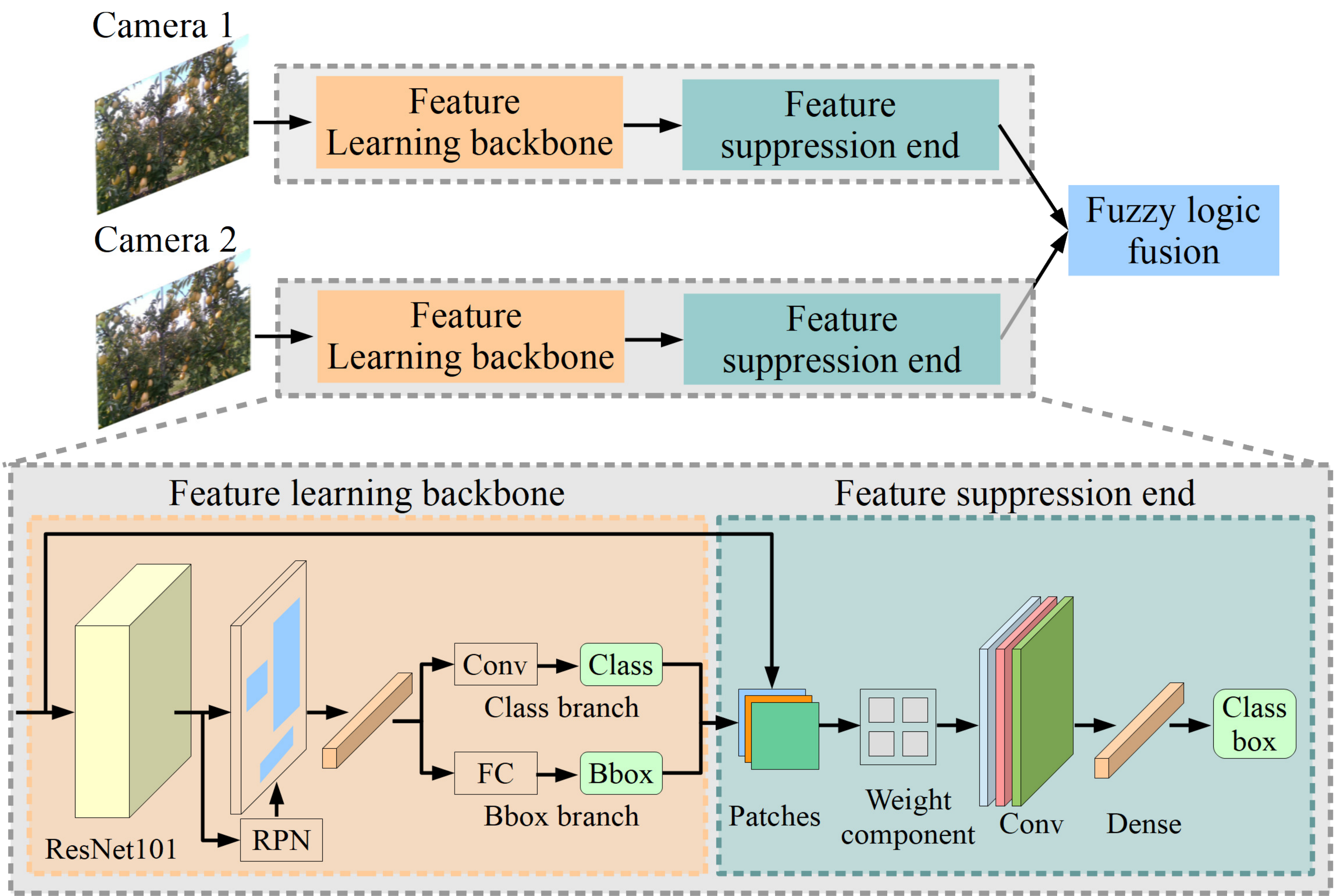}
		\caption{Apple detection structure based on two-camera setup.}\label{fig_appleDetStructure}
	\end{figure}
	
	\begin{figure}[!t]
		\centering
		\includegraphics[width=0.98\linewidth]{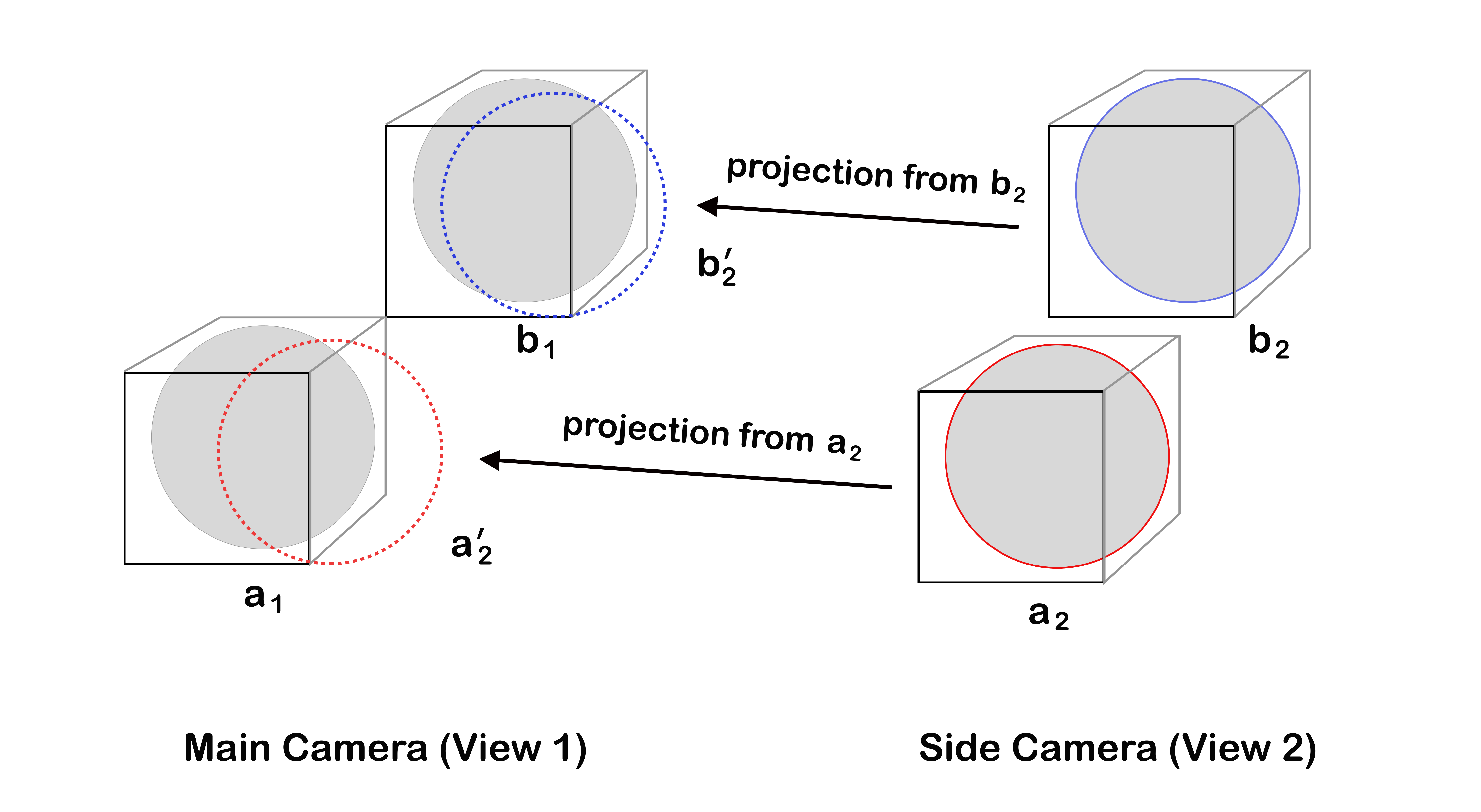}
		\caption{Apple matching unit of the fusion scheme: assume we have two apples $a$ and $b$ which are identified as $a_1$, $b_1$ in the main camera (View 1) and are identified as $a_2$, $b_2$ in the side camera (View 2). Based on the extrinsic calibration of the two cameras, $a_2$, $b_2$ in the side camera are first transformed into $a_2^\prime$, $b_2^\prime$ in the main camera. Then we match $a_1$ with $a_2^\prime$ and match $b_1$ with $b_2^\prime$ based on the overlap.}
		\label{fig_bbox_match}
	\end{figure}
	
\begin{figure}[!h]
		\centering
		\includegraphics[width=1\linewidth]{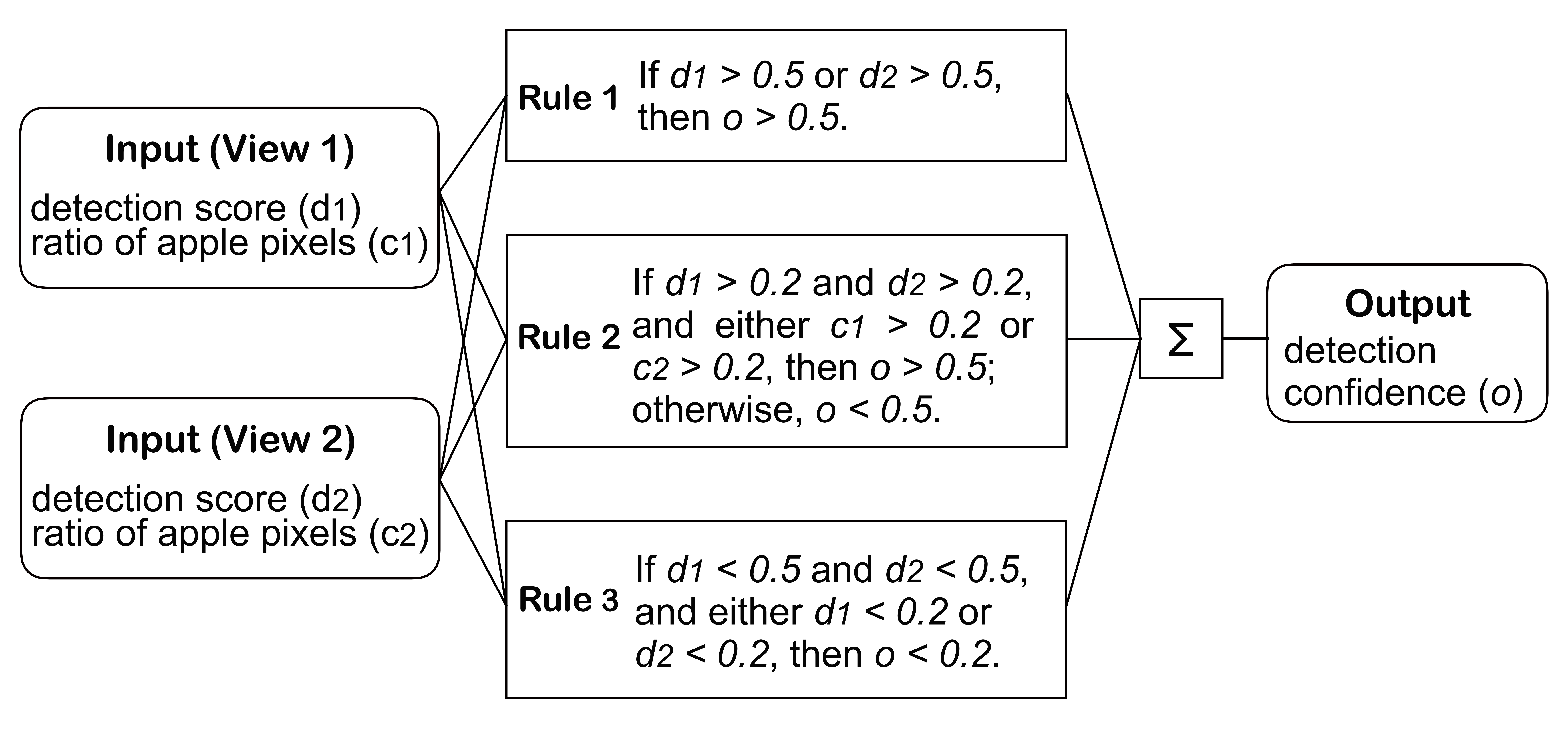}
		\caption{Fuzzy logic unit: the inputs are detection score ($d_1, d_2$) and the ratio of apple pixels ($c_1, c_2$) from both view 1 and view 2 with a limited range of $0-1$. We use fuzzy reasoning to evaluate all three rules in parallel and then the results of the rules are combined and distilled to the detection confidence $o$.}
		\label{fig_fuzzyLogic}
	\end{figure}

\begin{figure}[!h]
		\centering
		\includegraphics[width=8cm]{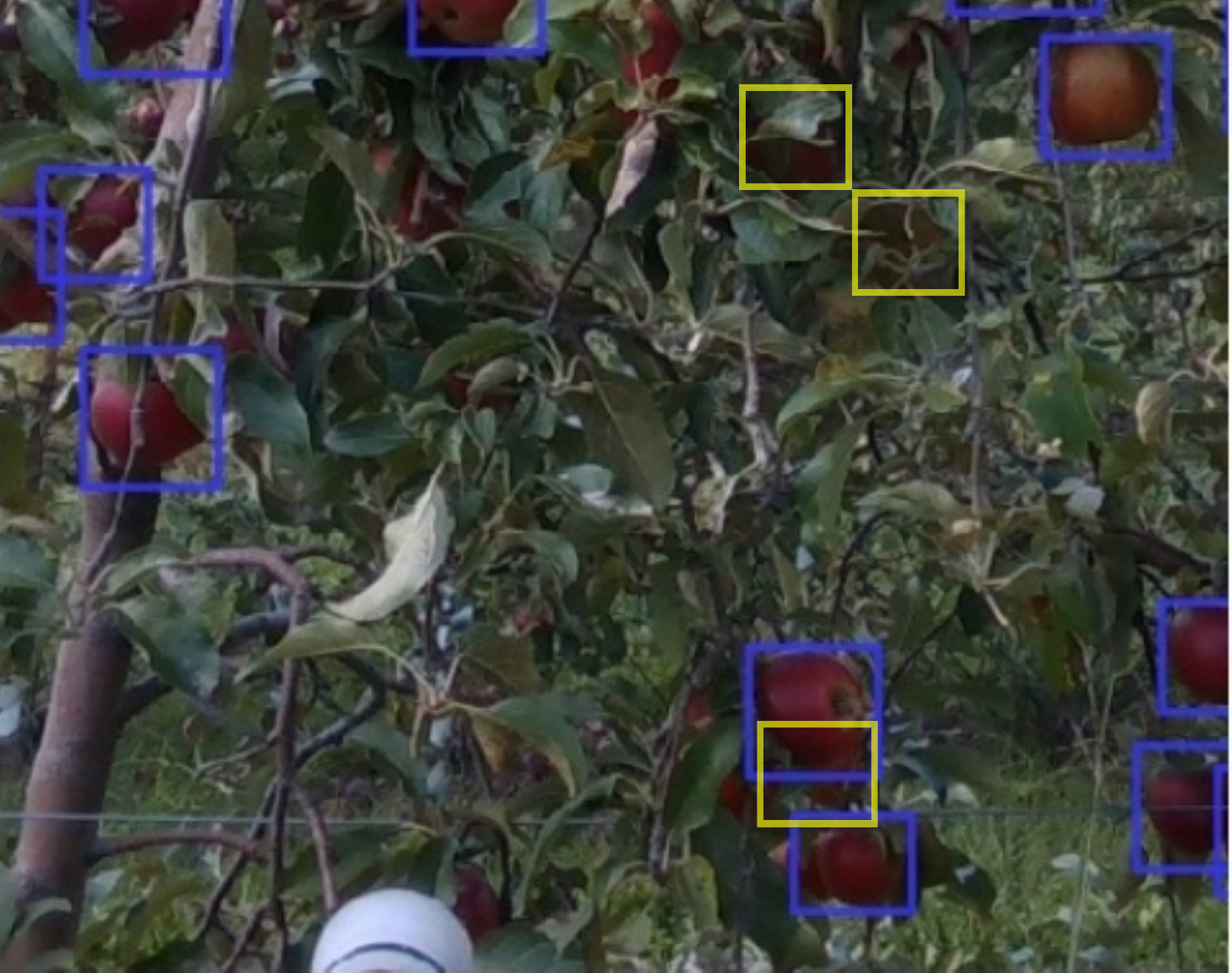}
		\caption{A comparison example between multi-view fusion detection and single-view detection. The blue bounding boxes represent the targets detected with the single-view detection algorithm. The yellow bounding boxes denote additional detection by introducing multi-view fusion mechanism.}\label{fig_fusionDetectionDemo}
	\end{figure}

	The network architecture of the proposed detection approach is shown in Fig. \ref{fig_appleDetStructure}. The raw images captured by the two RGB-D cameras from different perspectives are fed into identical but separate deep learning network which consists of two components: a feature learning backbone and a feature suppression end. The feature learning backbone adopts the classical backbone designed in Mask R-CNN \cite{he2017mask} to extract apple features and generate region proposals. 
	Since the feature learning backbone might generate wrong inference features, the image patches inside the bounding boxes are then passed to a feature suppression end to remove some mis-classified candidates. Once the images from the cameras are processed by the deep learning network, the bounding boxes of apple candidates are obtained. This suppression Mask R-CNN design has been reported in \cite{Chu2020PRL}. For the multi-view object detection, the key task is to associate the identical objects from two views. To merge the detection results from the two camera channels, we further design a fusion scheme that consists of an apple matching unit and a fuzzy logic unit. In the matching unit, we match the bounding boxes from the two camera frames based on the extrinsic calibration of the two cameras. As showed in Fig.~\ref{fig_bbox_match}, the bounding boxes and apple positions detected from the side camera are transformed into the corresponding ones represented under the coordinate frame of the main camera. Based on the overlap relationship, we build associations for apples between two cameras.
	For the matched and unmatched bounding boxes, a fuzzy logic unit is utilized to combine the detection results to further enhance the accuracy of the labeled candidates. As shown in Fig.~\ref{fig_fuzzyLogic}, we design expert rules for the fuzzy logic unit. A triangular fuzzy membership function \cite{pedrycz1994triangular} is applied to process the crisp input, and the results are distilled to determine the final detection confidence.  Fig.~\ref{fig_fusionDetectionDemo} shows an example of the detection results. It is clear that apples with high occlusion can be identified more effectively by introducing the multi-view fusion detection mechanism.
	Moreover, F1-score is used to quantitatively evaluate the detection performance. Specifically, all detection outcomes are divided into four types: true positive (TP), false positive (FP), true negative (TN), and
	false negative (FN), based on the relation between the true class and predicted class. Then precision (P) and recall (R) are defined as follows:
	\[
	\begin{aligned}
		P &=\frac{TP}{TP+FP},  
		\\
		R &= \frac{TP}{TP+FN}.
	\end{aligned}
	\]
	The F1-score is the harmonic mean of the precision and recall, which is defined as follows:
	\[
	F1=\frac{2P\cdot R}{P+R}.
	\]	
	The detection algorithm with two-camera setup achieves an F1-score of 93.92\%, while the one with a single camera achieves an F1-score of 90.5\%. The code on the network and fuzzy logic implementation is open sourced (\url{https://github.com/pengyuchu/DualCamFusion}).
	
	After detecting the fruits, apple localization is performed by employing the depth information provided by the RGB-D camera. More precisely, for each bounding box, the image pixels of the detected apple are extracted to generate a range matrix  by utilizing the disparity map. We then calculate the mean value of the range matrix and regard it as the apple's depth range. Combining the depth range with the center of the bounding box pixels, back-projection \cite{Hartley2003} is used to calculate the 3D position of the apple. This process is conducted for each of the bounding boxes to obtain positions of all detected apples in the image.
	
	\subsection{Unified picking/dropping and motion planning} \label{subsec_planning}
	In our system, there are two levels of planning tasks. At a high level, we need to plan the apple picking sequence and dropping spots based on the list of detected apple locations provided by the perception module. At a lower level, we need to  generate reference trajectory for a selected target apple for the manipulation control.
	
	The high-level harvesting sequence planning is necessary as it plays a crucial role in reducing the harvesting cycle time. Different from existing works \cite{silwal2017,Edan1991} that only focus on optimizing the fruit picking sequence, we take both the apple picking and dropping spot sequences into consideration. This flexibility is enabled by our dropping module design where the end effector does not need to return all the way to the home position to release the picked fruits. As shown in Fig.~1, the dropping module allows the end effector to release the detached apple in a large area, offering additional flexibility and optimization freedom for improved harvesting efficiency. More specifically, given $N$ detected apples with $p_{i}\in \mathbb{R}^{3}$, $i=1, \cdots, N$, denoting their 3D positions expressed in the manipulator frame. The apple picking sequence and the dropping spot sequence are defined as follows:
	\begin{itemize}
		\item The apple picking sequence $S= \left\lbrace s_{1}, \cdots, s_{N} \right\rbrace$ is a permutation of $ \left\lbrace 1, \cdots, N\right\rbrace$, which determines the sequence of picking apples  with the corresponding position sequence $\left\lbrace p_{s_{1}}, \cdots, p_{s_{N}} \right\rbrace$ that the manipulator will follow to travel through. Each apple only will be visited once.
		
		\item The dropping spot sequence $S_{d}= \left\lbrace \bar{p}_{s_{1}}, \cdots, \bar{p}_{s_{N-1}} \right\rbrace$ is a list of ordered 3D positions where the manipulator will stop by and release the harvested fruit. As discussed in Section~\ref{sub_hardware}, the dropping module provides a specific domain (which is denoted by $\bar{\mathbb{P}}$) for the manipulator to release the fruit, and hence the dropping spots $\bar{p}_{s_{i}}$ should be generated from this domain, i.e., $\bar{p}_{s_{i}}\in \bar{\mathbb{P}}$.  
	\end{itemize}
	In the planning phase, we consider that the manipulator will start from its home position $p_{0}$ and approach the detected apples by following the sequence $S$ defined above. For $i=1, \cdots, N-1$, once the apple located at $p_{s_{i}}$ is harvested, the manipulator will move to the position $\bar{p}_{s_{i}}$ to release the fruit and then heads to the next apple located at $p_{s_{i+1}}$. In particular, if the last apple in the picking sequence is harvested, the manipulator will return back to the home position $p_{0}$ for fruit release. According to the above description, it can be concluded that the manipulator's maneuver satisfies the following sequence:
	\begin{equation} \label{planning_seq}
		p_{0} \!\rightarrow\! p_{s_{1}} \!\rightarrow\! \bar{p}_{s_{1}} \!\rightarrow\! \cdots \!\rightarrow\! p_{s_{N-1}} \!\rightarrow\! \bar{p}_{s_{N-1}} \!\rightarrow\! p_{s_{N}} \!\rightarrow\! p_{0}.
	\end{equation}
	The planning objective is to determine the picking sequence $S$ and its corresponding dropping spot sequence $S_{d}$ by optimizing the travel cost along the maneuver sequence \eqref{planning_seq}. We use Euclidean distance to define the travel cost, as follows:
	\begin{equation} \label{eq_g}
		g = \| p_{s_{1}}-p_{0} \| + \sum_{i=1}^{N-1} g_{s_{i}, s_{i+1}} + \|p_{0}-p_{s_{N}} \|,
	\end{equation}
	where $g_{s_{i}, s_{i+1}} = \| \bar{p}_{s_{i}}-p_{s_{i}} \| + \|p_{s_{i+1}}-\bar{p}_{s_{i}}\| \in \mathbb{R}$ ($i=1, \cdots, N-1$) is the travel cost between two adjoining apples in $S$. Given $p_{s_{i}}$ and $p_{s_{s+i}}$, the optimal dropping spot $\bar{p}^{*}_{s_{i}}$ can be determined by solving the following problem:
	\begin{equation} \label{planning_opt1}
		\begin{aligned}
			&\min_{\bar{p}_{s_{i}}} g_{s_{i}, s_{i+1}}(\bar{p}_{s_{i}}) = \| \bar{p}_{s_{i}}-p_{s_{i}} \| + \|p_{s_{i+1}}-\bar{p}_{s_{i}}\|,
			\\
			&\text{s.t.} \quad \bar{p}_{s_{i}}\in \bar{\mathbb{P}}.
		\end{aligned}
	\end{equation}
	Let $g_{s_{i}, s_{i+1}}^{*} = \| \bar{p}_{s_{i}}^{*}-p_{s_{i}} \| + \|p_{s_{i+1}}-\bar{p}_{s_{i}}^{*}\| \in \mathbb{R}$ be the minimal value of $g_{s_{i}, s_{i+1}}$. Then, the optimization problem over the picking sequence $S$ is formulated as 
	\begin{equation} \label{planning_opt2}
		\begin{aligned}
			&\min_{S} g(S) = \| p_{s_{1}}-p_{0} \| + \sum_{i=1}^{N-1} g_{s_{i}, s_{i+1}}^{*} + \|p_{0}-p_{s_{N}} \|,
			\\
			&\text{s.t.} \quad s_{i} \in \left\lbrace 1, \cdots, N \right\rbrace, i=1, \cdots, N,
			\\
			& \qquad \, s_{i} \neq s_{j}, \text{for any}\; i\neq j\; \text{and}\; i, j = 1, \cdots N.
		\end{aligned}
	\end{equation}
	To determine the apple picking sequence and the dropping spot sequence. We first calculate the minimal travel cost between any two apple positions via \eqref{planning_opt1}. With the obtained minimal travel cost for any pair of two apples, the optimization problem \eqref{planning_opt2} can be reformulated as a travel salesman problem (TSP). The nearest neighbor algorithm~\cite{gutin2006TSP} is utilized to address the TSP, and then the apple picking sequence $S$ and the dropping spot sequence $S_{d}$ can be determined. 
	
	Based on the sequences $S$ and $S_{d}$, the apple location $p_{s_{i}}$ and the dropping spot $\bar{p}_{s_{i}}$ will be assigned in turn as the targeted position $p_{d}=\begin{bmatrix}
		x_{d}, y_{d}, z_{d}
	\end{bmatrix}^{\top}$ where the manipulator needs to reach. In our implementation, the planning algorithm described above will be performed whenever the perception system sends an updated list of detected apple locations. To facilitate the manipulation control, given a targeted position $p_{d}$ (e.g., top of the picking list), we use the quintic function \cite{siciliano2010robotics} to generate a corresponding reference trajectory $p_{r}(t)=\begin{bmatrix}
		x_{r}(t),\, y_{r}(t),\, z_{r}(t)
	\end{bmatrix}^{\top}$. This reference trajectory is a function of time with its terminus being the target position $p_{d}$. The introduction of the quintic function-based reference trajectory $p_{r}$ brings the following advantages: First, the reference trajectory is continuously differentiable and its terminal velocity and acceleration are zero, which is conducive to ensuring that the end-effector approaches the desired position along a smooth path. Second, by adjusting function parameters, the velocity profile of the reference trajectory can be modified, and thus the end-effector can reach the desired position within a specific time interval.
	
	\subsection{Efficient nonlinear control for accurate reference tracking} \label{subsec_control}
	Given a target apple and the generated reference trajectory using the planning algorithm discussed above, we next introduce the control algorithm that drives the manipulator to follow the reference trajectory.
	As shown in Fig. \ref{fig_flowDiagram}, one key requirement of the control module is to adjust the manipulator to approach the detected fruits or dropping spots with high accuracy and flexibility.
	To achieve this goal, a motion control strategy is developed by fully exploiting the mechanical structure of the developed 3-DOF manipulator.
	\begin{figure}[!b]
		\centering
		\includegraphics[width=8.8cm]{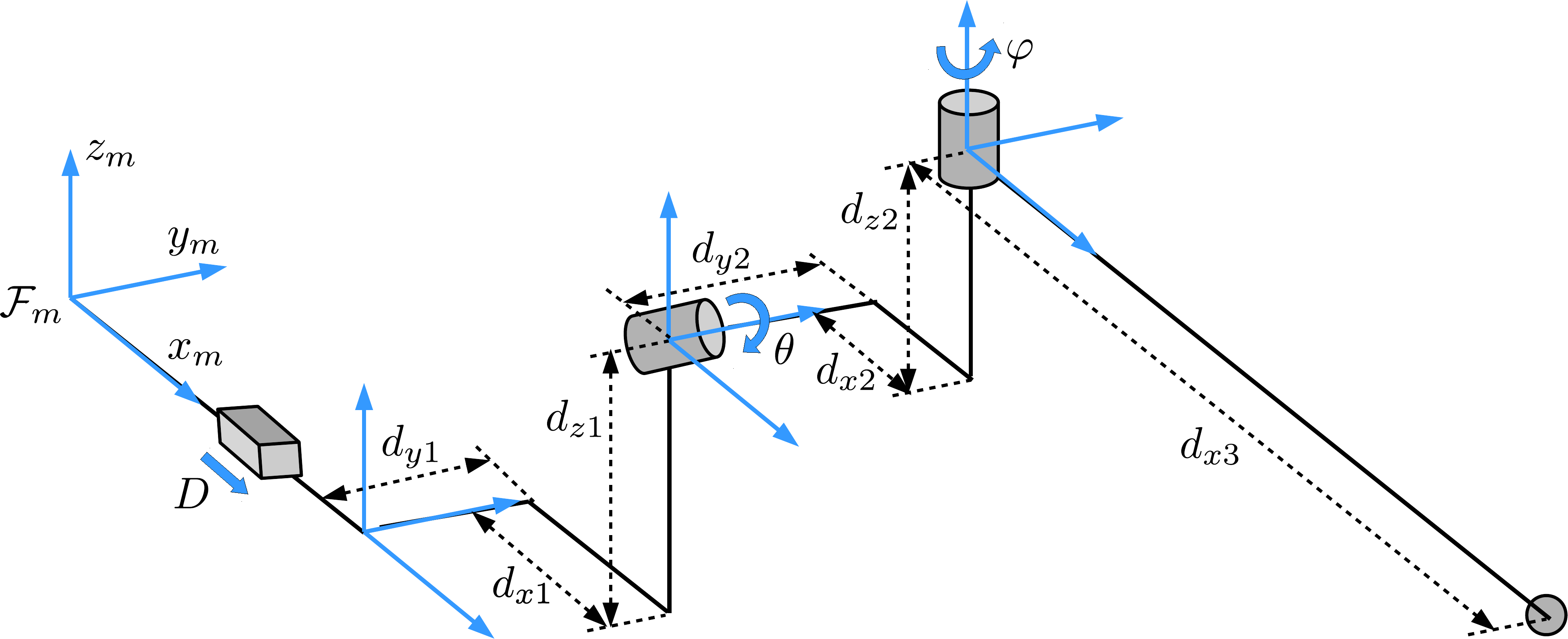}
		\caption{Kinematical description of the 3-DOF manipulator.}\label{fig_kineModel}
	\end{figure}
	
	The kinematic description of the 3-DOF manipulator is shown in Fig. \ref{fig_kineModel}. Denote $p = \begin{bmatrix}
		x, y, z
	\end{bmatrix}^{\top} \in 
	\mathbb{R}^{3}$ as the position of the end-effector. Based on the Denavit–Hartenberg convention \cite{siciliano2010robotics} and the kinematical diagram presented in Fig. \ref{fig_kineModel}, the forward kinematics function of the manipulator can be derived, as follows:
	\begin{equation} \label{eq_xyz}
		\begin{aligned}
			x &= d_{x3}\cos(\theta) \cos(\varphi) + d_{x2}\cos(\theta) + d_{z2}\sin(\theta) + d_{x1} + D,
			\\
			y &= d_{x3}\sin(\varphi) + d_{y2} + d_{y1},
			\\
			z &= -d_{x3}\sin(\theta) \cos(\varphi) - d_{x2}\sin(\theta) + d_{z2}\cos(\theta) + d_{z1},
		\end{aligned}
	\end{equation}
	where $d_{x1}$, $d_{x2}$, $d_{x3}$, $d_{y1}$, $d_{y2}$, $d_{z1}$, $d_{z2} \in \mathbb{R}$ are the link lengths, and $\begin{bmatrix}
		\varphi, \theta, D
	\end{bmatrix}^{\top} \in \mathbb{R}^{3}$ are the joint variables. 

	As described in Section \ref{subsec_planning}, the planning module will provide the reference trajectory $p_{r}(t)=\begin{bmatrix}
		x_{r}(t),\, y_{r}(t),\, z_{r}(t)
	\end{bmatrix}^{\top}$ for the targeted position $p_{d}$. The objective of the manipulation control is to regulate the end-effector to follow the reference trajectory $p_{r}$ and finally approach the target position $p_{d}$.
	The revolute joint parameters $\varphi$, $\theta$ and prismatic joint parameter $D$ are all driven by electrical motors, and the velocity-based control scheme is employed to generate explicit speed command to smoothly adjust the joints based on real-time position feedback.
	Specifically, based on \eqref{eq_xyz}, the time derivative of $\begin{bmatrix} x, y, z \end{bmatrix}^{\top}$ can be calculated as
	\begin{equation} \label{eq_dot_yz}
		\begin{aligned}
			\dot{x} &= -d_{x3}(\sin(\theta)\cos(\varphi)\omega_{\theta} + \cos(\theta)\sin(\varphi)\omega_{\varphi} ) 
			\\
			&\quad - d_{x2}\sin(\theta)\omega_{\theta} + d_{z2}\cos(\theta)\omega_{\theta} + v_{D},
			\\
			\dot{y} &= d_{x3}\cos(\varphi) \omega_{\varphi},
			\\
			\dot{z} &= -d_{x3}(\cos(\theta) \cos(\varphi) \omega_{\theta} - \sin(\theta)\sin(\varphi) \omega_{\varphi})
			\\
			&\quad -d_{x2}\cos(\theta)\omega_{\theta} - d_{z2}\sin(\theta)\omega_{\theta},
		\end{aligned}
	\end{equation}
	where $\omega_{\varphi}$, $\omega_{\theta} \in \mathbb{R}$ are the angular velocity of the revolute joints $\varphi$ and $\theta$, respectively, and $v_{D}\in \mathbb{R}$ is the linear velocity  of the prismatic joint $D$. Furthermore, the error signals $\begin{bmatrix} e_{x},
		e_{y}, e_{z}
	\end{bmatrix}^{\top} \in \mathbb{R}^{3}$ are constructed as
	\begin{equation} \label{eq_ey_ez}
		\begin{aligned}
			e_{x} &= x-x_{r},
			\\
			e_{y} &= y-y_{r}, 
			\\
			e_{z} &= z-z_{r}.
		\end{aligned}
	\end{equation}
	
	Based on \eqref{eq_dot_yz}, \eqref{eq_ey_ez}, and by virtue of Lyapunov-based control techniques \cite{khalil2002nonlinear}, the velocity controller is designed as
	\begin{equation} \label{eq_omega}
		\begin{aligned}
			\omega_{\varphi} & = \frac{-k_{y}e_{y} + \dot{y}_{r}}{d_{x3}\cos(\varphi)},
			\\
			\omega_{\theta} & = \frac{k_{z}e_{z} + d_{x3}\sin(\theta)\sin(\varphi)\omega_{\varphi} - \dot{z}_{r}}{d_{x3}\cos(\theta) \cos(\varphi) + d_{x2}\cos(\theta) + d_{z2}\sin(\theta)},
			\\
			v_{D} & = -k_{x}e_{x} +d_{x3}(\sin(\theta)\cos(\varphi)\omega_{\theta} + \cos(\theta)\sin(\varphi)\omega_{\varphi} ) 
			\\
			&\quad + d_{x2}\sin(\theta)\omega_{\theta} - d_{z2}\cos(\theta)\omega_{\theta} + \dot{x}_{r},
		\end{aligned}
	\end{equation}
	where $k_{x}$, $k_{y}$, $k_{z} \in \mathbb{R}^{+}$ are positive constant gains. 
	The velocity controller \eqref{eq_omega} can ensure that the end-effector position tracks the reference trajectory $\begin{bmatrix}
		x_{r}, y_{r}, z_{r}
	\end{bmatrix}^{\top}$ asymptotically, and rigorous stability analysis can be conducted based on the Lyapunov-based control techniques \cite{khalil2002nonlinear}.

	\section{Experiment and Results} \label{sec_perfEva}
	
	In this section, both indoor and field experiments are presented to demonstrate the performance of the developed system. The indoor tests with artificial trees are focused on validating the planning and control schemes, and the integrated system is further evaluated in the orchard field.
	
	\subsection{Indoor test validation}
	Indoor tests were carried out in a simulated orchard environment, which consists of artificial trees hanging real apples, a flat-panel lighting system, and a Qualisys motion capture system. 
	
	\begin{figure}[!h]
		\centering
		\includegraphics[width=8cm]{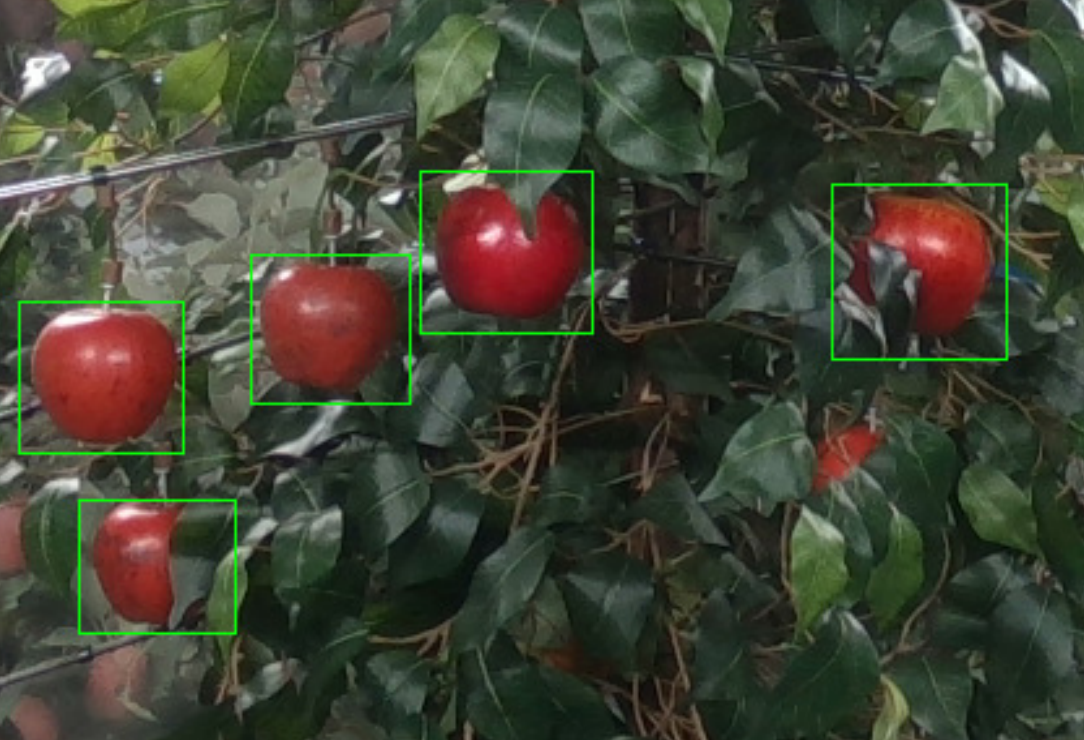}
		\caption{The scenario of Trial 1 for planning evaluation, where green bounding boxes represent detected apples.}\label{fig_scenePlanning}
	\end{figure}
	We first validate the performance of the planning algorithm that is used to generate picking sequence and dropping spots based on the apple locations provided by the perception module. 
	Three trials with different apple configurations are carried out for thorough validation. These three trials each have 5 apples, 7 apples, and 9 apples randomly hung on the artificial tree, and Fig. \ref{fig_scenePlanning} depicts the scenario of Trial 1 for reference.
	The perception algorithm is used to detect and localize these fruits, and then the planning module is triggered to determine the picking sequence and dropping spots. The travel cost defined in \eqref{eq_g} is calculated with the obtained picking sequence and dropping spots. We also actuate the manipulator to reach the apple positions and the dropping spots by following the planning results, and the total travel time (i.e., the maneuver time of the manipulator) is recorded. Moreover, to better demonstrate the effectiveness of the planning algorithm, the non-planning case is introduced for comparison. In the non-planning case, we do not optimize the dropping spots and consider that once the manipulator reaches a detected apple, it will always move to the home position to release the fruit. The same testing scenarios are used to obtain the travel cost and travel time under the non-planning case. The results are summarized in Table \ref{table planning}. It is clear that the proposed planning algorithm can significantly reduce the travel cost by optimizing the picking sequence and dropping spots. Furthermore, the travel time is an intuitive indicator to show the effect of the planning module in reducing the harvesting cycle time. Compared to the non-planning case, the proposed planning algorithm can efficiently reduce the travel time. 
	\begin{table}[!t]
		\caption{Comparison of travel distance and travel time  between the proposed planning scheme and non-planning case}
		\label{table planning}
		\begin{center}
				\begin{threeparttable}
					\begin{tabular}{c c c c}
						\hline
						\hline
						Trial & Fruit number & Travel distance NP (\textbf{P}) & Travel time NP (\textbf{P}) \\
						\hline
						1 & 5 & 4.12 (\textbf{3.18}) [m] & 12.55 (\textbf{10.70}) [s] \\  
						2 & 7 & 5.61 (\textbf{4.28}) [m] & 17.11 (\textbf{14.51}) [s] \\
						3 & 9 & 6.81 (\textbf{4.86}) [m] & 21.07 (\textbf{17.65}) [s] \\
						\hline
						\hline
					\end{tabular}
					\begin{tablenotes}
						\footnotesize
						\item where NP $=$ non-planning case and P $=$ proposed planning scheme.
					\end{tablenotes}
				\end{threeparttable}
		\end{center}
	\end{table}
	
	To thoroughly validate the control performance, the manipulator are driven to multiple target positions, and then the Qualisys motion capture system is utilized to evaluate the motion accuracy. Specifically, as discussed in Section \ref{subsec_control}, the developed manipulator includes three joints, i.e., $\begin{bmatrix}
		\varphi, \theta, D
	\end{bmatrix}^{\top}$. The desired joint values are selected from the following sets: $\varphi_{d}, \theta_{d} \in \left\lbrace -20^{\circ}, -10^{\circ}, 10^{\circ}, 20^{\circ} \right\rbrace$ and $D_{d} \in \left\lbrace 0.1 \text{m}, 0.3 \text{m}, 0.5 \text{m} \right\rbrace$,
	and the corresponding target positions can be computed based on \eqref{eq_xyz}. A total of 48 target positions are generated, which are evenly distributed in the workspace of the manipulator. Furthermore, a spherical marker is attached to the end-effector, ensuring that the Qualisys motion capture system can measure the end-effector position precisely through marker identification. The manipulator is actuated from the home position to each of the target positions, and the final position of the end-effector is recorded. Based on 48 pairs of final position records and the corresponding target positions, the average errors along the $x$-axis, $y$-axis, $z$-axis and the average distance errors are calculated. The controller designed in \eqref{eq_omega} is tested, and the results are shown in Table \ref{table distance error}. It can be seen that the average distance error is less than 1 cm, indicating that the proposed control scheme can achieve satisfactory performance.
	\begin{table}[!t]
		\caption{Average absolute error between target positions and manipulator final positions with the designed controller}
		\label{table distance error}
		\begin{center}
				\begin{tabular}{c c}
					\hline
					\hline
					& Nonlinear controller \eqref{eq_omega} \\
					\hline
					$x$-axis error [cm]
					& 0.3905 \\ 				
					$y$-axis error [cm]
					& 0.3324 \\
					$z$-axis error [cm]
					& 0.2742 \\
					Distance error [cm]
					& 0.6566 \\            
					\hline
					\hline
				\end{tabular}
		\end{center}
	\end{table}
	
	\subsection{Field test and validation}
	To further evaluate the performance of the integrated prototype, field experiments are conducted in the Horticultural Teaching and Research Center of Michigan State University during the 2021 harvest season\footnote{A video of the robotic apple harvester demonstrating the field tests is available at \url{https://www.youtube.com/watch?v=\_6-5qbZplZo}}. 
	
	In the field test, the robotic apple harvesting system is run autonomously and continuously to harvest fruits within its workspace with fully integrated perception, planning, and control functionalities.
	On average, the duration for the manipulator to approach a target apple or move back to its corresponding dropping spot ranges between 0.75 s and 1.4 s. Detaching and releasing the fruit roughly take 1 s and 0.5 s, respectively. For the successfully harvested apples, the average cycle time is approximately 3.6 s, including software algorithm processing and hardware execution. Compared to our previous prototype \cite{Zhang2021MECH} and the existing literature \cite{baeten2008,silwal2017,Hohimer2019,Bulanon2021} which took 7-10 seconds to harvest an apple, the current robotic apple harvesting prototype clearly has made significant advancement in terms of harvesting efficiency, thanks to the simple yet efficient mechanism  as well as the integrated algorithm design. 
	
\begin{figure}[!t]
		\centering
		\subfigure[] {\label{img_failedCase1}
			\includegraphics[width=0.225\textwidth]{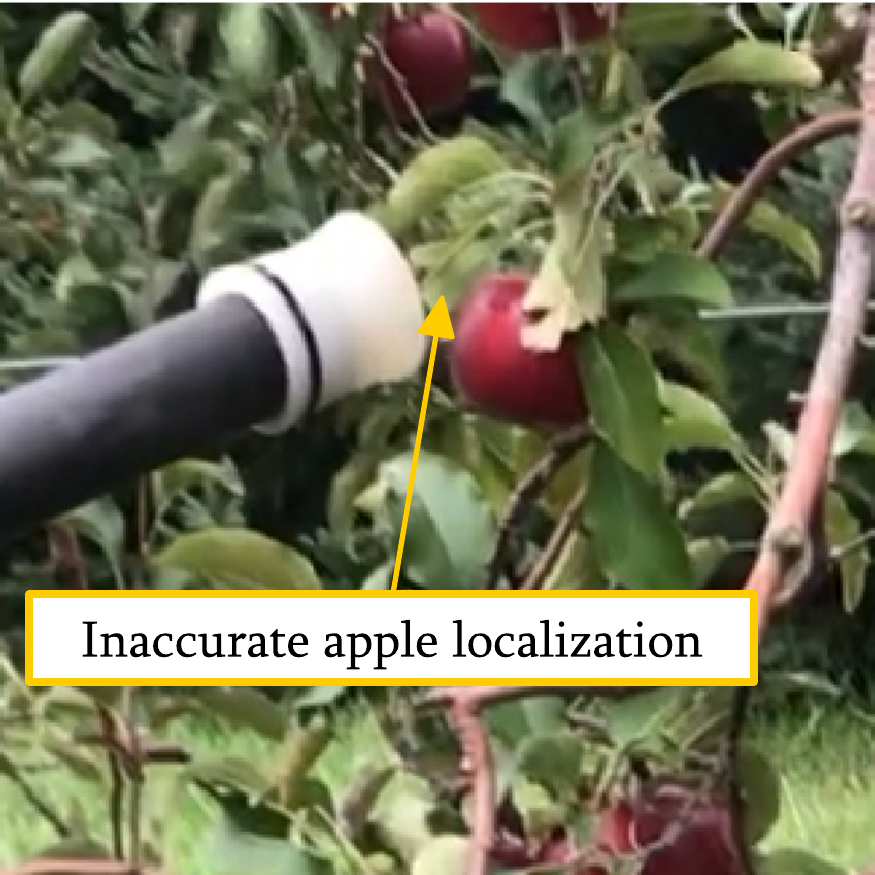}
		}
		\subfigure[] {\label{img_filedCase2}
			\includegraphics[width=0.225\textwidth]{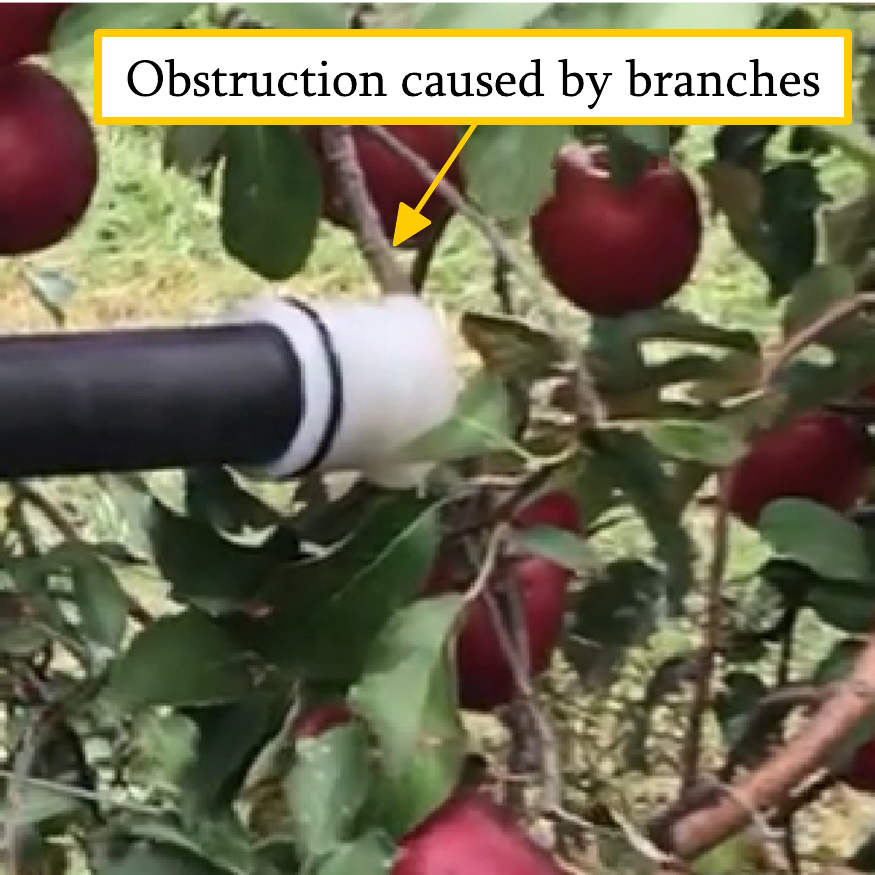}
		}
		\caption{Examples of failed apple harvesting: (a) due to inaccurate localization; and (b) due to obstruction caused by branches.}
		\label{fig_failedCase}
	\end{figure}

	However, there is still a considerable gap in achieving a satisfactory picking rate. In this field test, a total of 142 apples were attempted  and 74 of them were picked successfully with the picking rate being 52.1\%. We studied the failed cases and identified the following major causes. First, we found that the depth measurement in the RealSense RGB-D camera is susceptible to varying lighting conditions and cases when fruit is partially obscured by foliage or branches, and sometimes provides inaccurate depth information with an error as large as 10 cm. Second, unlike the v-trellis structured orchard used in \cite{silwal2017} where most of the apples are well exposed, the orchard where we conducted the experiment does not have a well-structured fruiting system and a high percentage of fruits are being occluded by leaves and branches, which create challenges for the robotic system to approach the target fruits.
	Third, there are also many occurrences during which the end-effector has failed to detach target fruits due to inadequate vacuum power. Fig. \ref{fig_failedCase} illustrates two failed harvesting scenarios.
	
	These findings provide useful insights towards further improvement of our system. The two-camera setup enhances fruit detection and localization in the indoor environment but appears to have limited contribution to improving the accuracy of fruit localization in the field. Fusing additional sensing modalities such as LiDAR could be a good way to achieve robust fruit localization and is currently under investigation. We also need to design an object segmentation algorithm to identify the obstacles (trunks, branches) and develop a path planning scheme to avoid obstruction. Finally, the vacuum system and fruit detachment strategy should also be further improved for reliable fruit picking. 
	
	\section{Conclusion} \label{sec_conclusion}
	The algorithm design and integration for a newly-developed robotic apple harvesting prototype was introduced in the paper. The algorithm component is comprised of three core modules: perception, planning, and control. 
	Indoor and field experiments demonstrated that the developed algorithm component can synergistically work with the hardware component to achieve the primary apple harvesting functionalities, offering a promising picking cycle time of 3.6 seconds.
	Guided by lessons learned from these experiments, future work will include improving fruit localization accuracy and robustness, developing object segmentation algorithms for obstacle detection, and designing optimal path planning scheme for obstacle avoidance.

	\balance	
	\bibliographystyle{IEEEtran}
	\bibliography{IEEEabrv,reference}
\end{document}